\documentclass[10pt,twocolumn,letterpaper]{article}

\usepackage{cvpr}
\usepackage{times}
\usepackage{epsfig}
\usepackage{graphicx}
\usepackage{amsmath}
\usepackage{amssymb}
\usepackage{mmstyles}
\usepackage{booktabs}
\usepackage{multirow}
\usepackage{array}

\newcolumntype{P}[1]{>{\centering\arraybackslash}p{#1}}
\newcolumntype{M}[1]{>{\centering\arraybackslash}m{#1}}

\usepackage[pagebackref=true,breaklinks=true,letterpaper=true,colorlinks,bookmarks=false]{hyperref}

\cvprfinalcopy % *** Uncomment this line for the final submission

\def\cvprPaperID{222} % *** Enter the CVPR Paper ID here

\ifcvprfinal\fi
\begin{document}

%%%%%%%%% TITLE
\title{Unsupervised Bi-directional Flow-based Video Generation from one Snapshot}

% \and
\author{Lu Sheng$^{1,4}$,~~~Junting Pan$^2$,~~~Jiaming Guo$^3$,~~~Jing Shao$^2$,~~~Xiaogang Wang$^4$,~~~Chen Change Loy$^5$\\
$^1$School of Software, Beihang University~~~~~$^2$SenseTime Group Limited\\
$^3$Institute of Computing Technology, Chinese Academy of Sciences \\
$^4$CUHK-SenseTime Joint Lab, The Chinese University of Hong Kong \\
$^5$SenseTime-NTU Joint AI Research Centre, Nanyang Technological University\\
{\tt\small lsheng@ee.cuhk.edu.hk}}

\maketitle
%\thispagestyle{empty}

%%%%%%%%% ABSTRACT
\begin{abstract}
Imagining multiple consecutive frames given one single snapshot is challenging, since it is difficult to simultaneously predict diverse motions from a single image and faithfully generate novel frames without visual distortions.
In this work, we leverage an unsupervised variational model to learn rich motion patterns in the form of long-term bi-directional flow fields, and apply the predicted flows to generate high-quality video sequences.
In contrast to the state-of-the art approach, our method does not require external flow supervisions for learning. This is achieved through a novel module that performs bi-directional flows prediction from a single image. %The flows are learned to preserve spatiotemporal and structural coherence with the help of cycle flow consistency.%, and present content awareness by compositional fusion with the reference image.
In addition, with the bi-directional flow consistency check, our method can handle occlusion and warping artifacts in a principle manner.
%
%The proposed frame synthesis is not only a flow-based warping operation, it is also a trainable module that completes novel frames with hallucinated novel contents.
%
%Our \emph{ImagineFlow} model offers a more natural way to describe motions and avoids directly processing in high-dimensional intensity space.
%
%It also eliminates inherent challenges in flow-based synthesis, such as occlusions and warping duplicates.
%
Our method can be trained end-to-end based on arbitrarily sampled natural video clips, and it is able to capture multi-modal motion uncertainty and synthesizes photo-realistic novel sequences.
Quantitative and qualitative evaluations over synthetic and real-world datasets demonstrate the effectiveness of the proposed approach over the state-of-the-art methods.\footnote{This work was done when Lu Sheng was with the CUHK-Sensetime Joint Lab, the Chinese University of Hong Kong.}
\end{abstract}

\section{Introduction}
\label{sec:introduction}

% ==== 1st: background/challenge of video generation from a single snapshot

We wish to address the problem of training a deep generation model for imagining photorealistic videos from just a \emph{single} image. 
The problem is a non-trivial task as the model cannot only guess plausible \emph{dynamics} conditioned on \emph{static contents}.
The problem is thus significantly harder than motion estimation task or video prediction problem in which paired consecutive images are assumed.
Under the single-image constraint, choosing a suitable representation learning method becomes critical to the final visual quality and plausibility of the rendered novel sequences.

% ==== 2nd: motion representation motivation, benef

\begin{figure}[t]
\centering
\includegraphics[width=\linewidth]{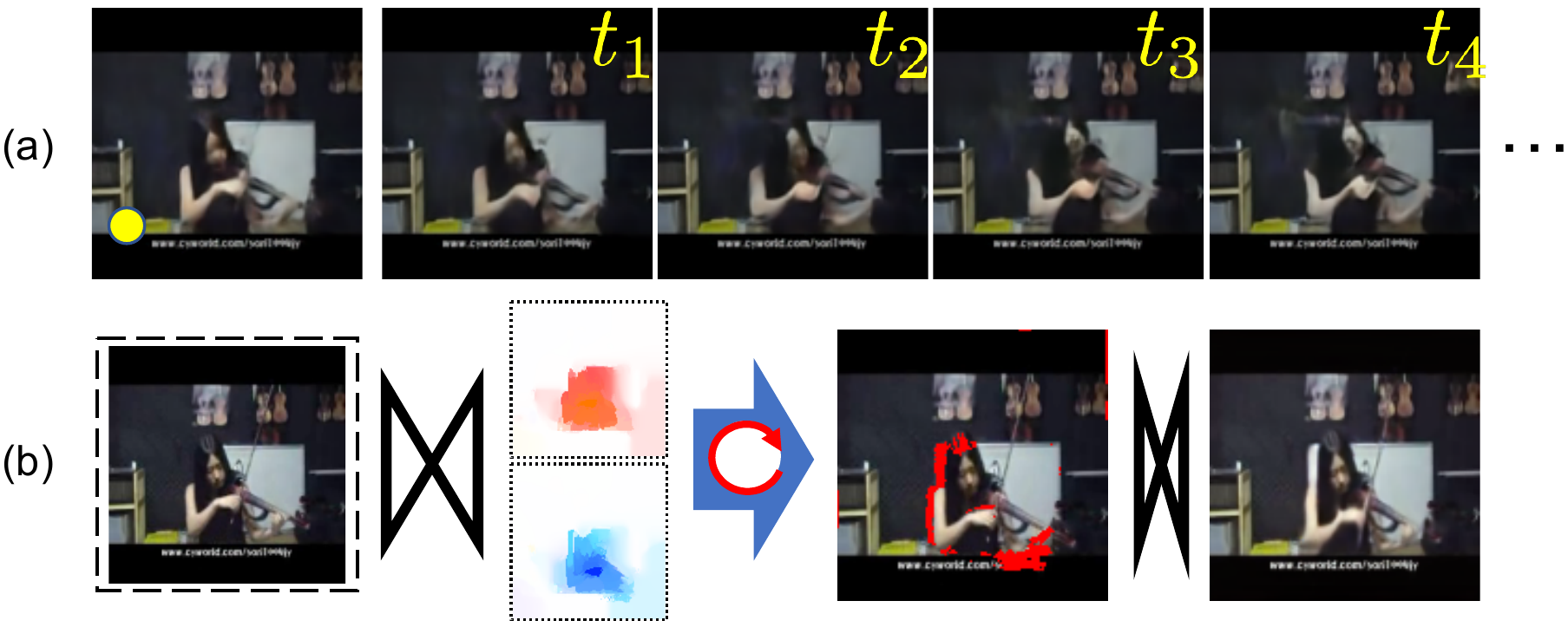}
\caption{{(a) Li~\etal~\cite{li2018flow} applies flows for single image-based video generation. But it cannot explicitly handle occlusions and its warping artifacts would be accumulated and the rendered frames are progressively degraded. Please see how the violinist is distorted over time. (b) The proposed ImagineFlow generates bi-directional flows, which can self-regularize the flow distribution and in principle tackle occlusions/warping artifacts. The rendered frame could create reasonable background that was once occluded.}}
\label{fig:intro}
\end{figure}

Existing methods such as MoCoGAN~\cite{tulyakov2017mocogan}, VGAN~\cite{vondrick2016generating}, Visual Dynamics~\cite{xue2016visual}, and FRGAN~\cite{zhao2018learning} either directly render RGB \emph{pixel values} or \emph{residual images} for the modeling of dynamics in novel sequences, but they usually distort appearance patterns and are thus short for preserving visual quality.
A recent method proposed by Li~\etal~\cite{li2018flow} learned dense flows to propagate pixels from the reference image directly to the novel sequences, offering a better chance to generate visually plausible videos.
However, Li~\etal~\cite{li2018flow} require external and accurate flow supervisions (generated from SpyNet~\cite{zhou2016view} to train the flow generation component, and the inherent warping artifacts were not handled in a principle way.
Warping artifacts frequently occur in the generated novel frames, such as the examples in Fig.~\ref{fig:intro}.

In this paper, we adopt the same notion of `per-pixel flow propagation' as in Li~\etal~\cite{li2018flow} but with the following consideration:
(1) how to learn robust content-aware flow distributions in an unsupervised manner, without any external flow supervision; 
(2) how to synthesize photorealistic frames while eliminating warping artifacts such as occlusions and warping duplicates in a principle manner.

To this end, we propose an unsupervised bi-directional flow-based video generation framework solely based on one single image, named as {ImagineFlow}, which tackles the aforementioned challenges in a unified way.
Our model has three appealing properties:

\vspace{+1mm}
\noindent
(i) \emph{End-to-End Unsupervised Learning} -- 
It allows end-to-end unsupervised learning to generate novel video from a single snapshot based on per-pixel flow propagation. The unsupervised learning module is new in the literature. It relaxes the need of external flows for supervision.
The proposed learning is incredibly convenient and powerful in our task as it does not require laborious ground-truth annotations or preparation and the training data is nearly infinite and rich in motion patterns. 

\vspace{+1mm}
\noindent
(ii) \emph{Bi-directional Flow Generation} -- 
We formulate a bi-directional flow generator, which outputs \emph{forward flows} (input image $\rightarrow$ target images), and the \emph{backward flows} (target frames $\rightarrow$ input image) simultaneously.
The bi-directional flows can self-regularize each other according to a well-known cycle consistency criteria, \ie, valid flows always have corresponding inverse flows back to their original locations.
The resultant flow distributions are within reasonable flow manifolds even they are learned without explicit flow supervision.
In contrast, Li~\etal~\cite{li2018flow} only generate the backward flows, whose reliability is governed by explicit flow supervisions.

\vspace{+1mm}
\noindent
(iii) \emph{Occlusion-aware Image Synthesis} --
Another merit of the proposed bi-directional flows generation is that the cycle consistency of flows allows us to detect occlusions in a robust and principle way.
Occlusions can be detected in areas where cycle consistency is violated.
Our model can leverage the occlusions inferred to help determine between per-pixel flow propagation or pixel hallucination for novel frame generation.
Specifically, our occlusion-aware image synthesis inputs bilinear warped~\cite{zhou2016view} novel frames overlaid with corresponded visibility (\ie, non-occluded) masks, and then employs a learnable mapping module that projects the warped frames onto the space of natural images.
This module generates reasonable contents in the disoccluded area and seamlessly repairs warping duplicates simultaneously.

\vspace{+1mm}
Ablation study validates the effectiveness of our {ImagineFlow} model. Extensive experimental evaluations also demonstrate its superior qualitative and quantitative performance over state-of-the-art methods~\cite{xue2016visual,vondrick2016generating,chen2017video,tulyakov2017mocogan,li2018flow}.

\section{Related Work}
\label{sec:related_work}

\noindent\textbf{Motion Prediction.}
Single image based video generation is closely related to the motion prediction problem.
Given an observed image or a short video clip, various methods have been proposed to predict dense future motions by optical flows~\cite{PinteaECCV2014,walker2015dense,walker2016uncertain}, object trajectories~\cite{walker2014patch}, difference or residual images~\cite{xue2016visual}, or deep visual representations~\cite{vondrick2016anticipating}.
While most methods follow a deterministic manner, a few seminal works also try to characterize the uncertainties in the predicted motions, based on probabilistic models such as conditional variational autoencoders~\cite{xue2016visual,walker2016uncertain} or generative adversarial networks~\cite{liang2017dual}.
Our work falls into a probabilistic motion modeling framework. We differ to aforementioned studies in that we aim at employing the predicted motion distribution to synthesize structurally coherent novel frames with reasonable motions. This requires new formulation for occlusion reasoning and frame generation.

\vspace{0.1cm}
\noindent\textbf{Motion-based Video Generation~}
Video generation can be roughly categorized into two classes according to whether it takes condition or not.
A series of \emph{unconditioned video generation} methods synthesize novel videos from scratch, with different learning techniques such as adversarial learning, motion/content separation, recurrent neural networks~\cite{srivastava2015unsupervised,tulyakov2017mocogan,saito2017temporal}.
The visual qualities of their outputs are still not satisfactory to generate photorealistic videos.
\emph{Conditioned video generation} appears to be a more promising choice to generate visually plausible videos.
A category of approaches predict novel frames from consecutive input frames~\cite{kalchbrenner2016video,srivastava2015unsupervised,patraucean2015spatio,luo2017unsupervised,mathieu2015deep,finn2016unsupervised}.
%
% Mathieu~\etal~\cite{mathieu2015deep} improved the details of the future frames by an adversarial loss and a unique gradient based loss.
% %
% Finn~\etal~\cite{finn2016unsupervised} uses differentiable motion predictive models to effectively sample multi-modal distributions of the predicted videos, but the results are still blurry and the model is limited to over-constrained scenarios.
%
Another category, which matches our problem setting, predicts future frames just based on one still image~\cite{zhao2018learning,li2018flow,chen2017video,xue2016visual}.
Some single image based methods characterize motions as feature filters or middle-level transformations~\cite{chen2017video,vondrick2017generating,xue2016visual}. These methods usually fail to preserve appearance patterns.
To achieve photorealistic generation quality, Li~\etal~\cite{li2018flow} also apply flows as the medium, but the method requires ground-truth flow supervisions and no explicit occlusion handling is introduced.
By contrast, our approach is unsupervised, as it learns to generate bi-directional flows directly through the constrain of cycle consistency. The flows are reliable even without explicit supervisions, and they permit occlusion detection in a principle manner.
With the structurally coherent flow fields, our method is more effective in rendering realistic video clips. Flow-based warping artifacts are eliminated by an additional occlusion-aware synthesis module.

\vspace{0.1cm}
\noindent\textbf{Applications and Regularizations for Flows.}
Flows have been adopted for various tasks, such as video enhancement by task-oriented flow~\cite{xue2017video}, video interpolation and extrapolation by voxel flow~\cite{liu2017video}, gaze direction detection~\cite{ganin2016deepwarp} and novel view synthesis~\cite{zhou2016view}.
Reliable flows often require regularizations to strengthen its structural coherence.
Some approaches regularize the structures explicitly by flow supervisions~\cite{dosovitskiy2015flownet,li2018flow} or implicitly by adversarial networks~\cite{liang2017dual}.
In this work, inspired by the cross validation in stereo and optical flow estimations, we found that simply with cycle flow consistency, the learned flow space can already be effectively enforced with structural coherence without laborious human labeling or preprocessing. 

\section{Methodology}
\label{sec:methodology}

\subsection{Problem Definition}
\label{sub:problem_definition}

Our task is to learn a probabilistic distribution $p(\mathcal{I}_\mathcal{T}|\mathbf{I}_0)$ conditioned on a reference frame $\mathbf{I}_0$, and then sample novel sequences $\tilde{\mathcal{I}}_\mathcal{T}$ from this conditioned distribution, where $\mathcal{T} = \{1,\ldots, T\}$ are the frame indices.

In this work, we aim at simultaneously predicting a set of backward flows $\mathcal{W}_\mathcal{T}^b = \{ \mathbf{W}_t^b\}_{t \in \mathcal{T}}$ pointing from tentative novel frames $\tilde{\mathcal{I}}_\mathcal{T}$ to the reference frame $\mI_0$, and corresponding forward flows $\mathcal{W}_\mathcal{T}^f = \{\mathbf{W}_t^f\}_{t \in \mathcal{T}}$ inversely from $\mI_0$ to $\tilde{\mathcal{I}}_\mathcal{T}$, and then leveraging the predicted bi-directional flow fields to generate novel sequences $\tilde{\cI}_{\cT}$.

More specifically, this task is equivalent to learning 1) a bi-directional flow generator $p_{\vphi}(\cW_\cT^b, \cW_\cT^f|\vz, \mI_0)$ that is also conditioned on a motion code $\vz$, and 2) an occlusion-aware image synthesis module $R_{\vomega}(\cdot)$.
The random motion code $\vz$ may be sampled from a standard Gaussian distribution.
$\vphi$ and $\vomega$ are network parameters.
The complete model is learned from a training set of sequence pairs $\{(\cI_\cT^{(n)}, \mI_0^{(n)})\}_{n = 1, \ldots, N}$ under an unsupervised manner.
%
% Note that our model is within the category of deep conditioned generative models~\cite{sohn2015learning,radford2015unsupervised}.

\subsection{Recap: Image Synthesis by Backward Warping}
\label{sub:naive_framework}

\begin{figure}[t]
\centering
\includegraphics[width=0.9\linewidth]{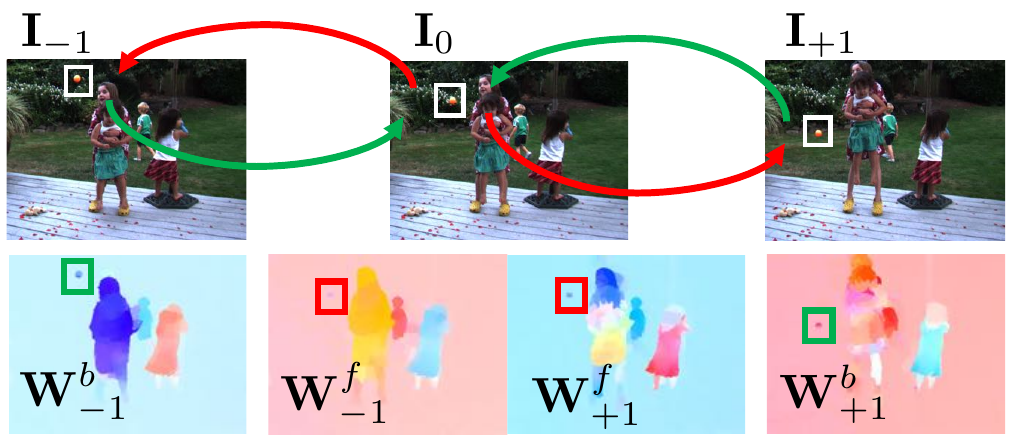}
\caption{Motion consistency in valid bi-directional flows. The red arrows indicate forward flows and green arrows are backward flows. Best viewed in screen.}
\label{fig:flow_cycle_consistency}
\end{figure}

Given the backward flows $\cW_\cT^b$ from the tentative target frames to the reference frame $\mI_0$, the synthesized frames are usually warped via bilinear sampling from $\mathbf{I}_0$~\cite{jaderberg2015spatial}:
\begin{equation}
\mI_{t\leftarrow0} = \mathcal{F}(\vx, \mW_t^b|\mI_0) = \mathbf{I}_0(\mathbf{W}_t^b(\mathbf{x}) + \mathbf{x}), ~\forall t \in \mathcal{T}.
\label{eq:backward_warping}
\end{equation}
Existing flow-based models, such as Li~\etal~\cite{li2018flow}, only apply the backward flows to generate novel frames.
However, in the framework of unsupervised learning, using backward flows alone is less favored due to three issues:

\vspace{0.1cm}
% \noindent\textbf{Occlusions and Warping Duplicates~}
\noindent\textbf{Warping Artifacts}.
Backward warping operations will produce artifacts due to occlusions.
Occlusions result in unfilled holes and generate warping duplicates in the warped images.

\vspace{0.1cm}
\noindent\textbf{Motion Inconsistency}.
Since unsupervised backward flow learning usually applies photometric consistency to learn the flow space, the learned backward flow distributions may not align with the real flow space, especially when the sequences contain plain area or repeated patterns.
Reliable flows should be cross-consistent outside the occlusion regions, \ie, an object in the target frames should always be able to predict reliable inverse flows back to the object in the reference frame, as illustrated in Fig.~\ref{fig:flow_cycle_consistency}, similar to the concerns arised in optical flow and stereo estimation~\cite{forsyth2002computer}.

\vspace{0.1cm}
\noindent\textbf{Structure Inconsistency}.
Moreover, $\mathbf{W}_t^b$ and novel frames (not the reference frame) are co-aligned in their spatial distributions, as shown in Fig.~\ref{fig:flow_cycle_consistency}.
It means that the backward flows do not only have to capture pixel-wise motions but also need to present the spatial structures in the target frames.
Unfortunately, na\"ive unsupervised learning usually tends to generate flows aligned with the condition $\mathbf{I}_0$ rather than novel sequences, thus neither the pixel-wise motion nor the spatial alignment can be well discovered.

\subsection{Bi-directional Flow Generation}
\label{sub:bi_directional_flow_generation}

Different from the backward flows, the forward flows $\mathcal{W}_\mathcal{T}^f$ are otherwise consistent with the spatial structure of $\mI_0$, as shown in Fig.~\ref{fig:flow_cycle_consistency}.
And ideally $\mathcal{W}_\mathcal{T}^f$ and $\mathcal{W}_\mathcal{T}^b$ should be cross-consistent except the occlusion regions.
Thus we also learn the forward flows $\mathcal{W}_\mathcal{T}^f$ as an auxiliary output concurrently with the backward flows $\mathcal{W}_\mathcal{T}^b$, and exploit these flows to regularize the spatial structures and enforce motion consistency of the backward flows.
Moreover, the cross-consistency between the bi-directional flows also give cues for the occlusion detection.

\begin{figure}[t]
\centering
\includegraphics[width=1\linewidth]{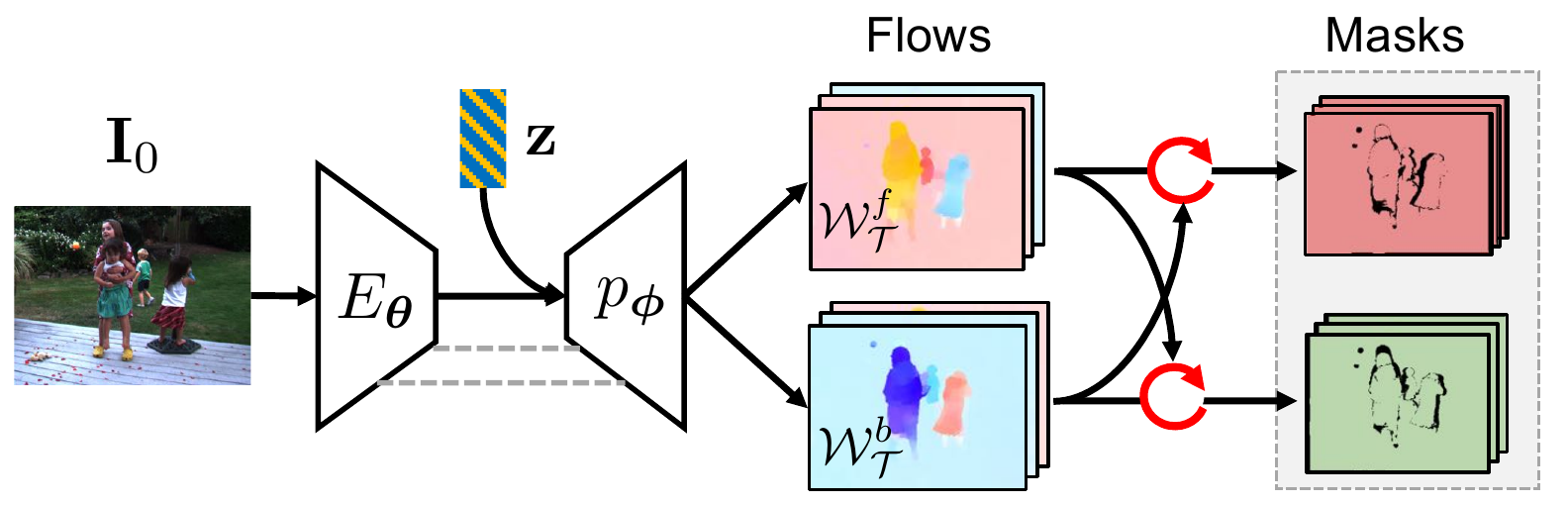}
\caption{Bi-directional flow generation. This network outputs bi-directional flows and further their visibility masks by cycle consistency. Content features from $E_{\boldsymbol\theta}(\mathbf{I}_0)$ are compositionally fused into the flow generator.}
\label{fig:circular_visual_consistency}
\end{figure}

We generate bi-directional flows from a bi-directional flow generator $p_{\boldsymbol\phi}(\mathcal{W}_\mathcal{T}^f, \mathcal{W}_\mathcal{T}^b | \vz, \mathbf{I}_0)$, with two parallel output branches, as visualized in Fig.~\ref{fig:circular_visual_consistency}.
The paired flows are constrained by the cross consistency that \emph{valid} pixel-wise paths by $\mathcal{W}_\mathcal{T}^b$ and $\mathcal{W}_\mathcal{T}^f$ form loop closure and bi-directional visual consistency between a training pair $\mathbf{I}_0$ and $\mathcal{I}_\mathcal{T} = \{ \mI_t \}_{t\in\cT}$.

\vspace{+1mm}
\noindent\textbf{Occlusion Detection~}
Occluded regions are usually the regions where the bi-directional flows are inconsistent.
We define the visibility mask $\mM_{0\leftarrow t}$ indicating pixels in $\mI_0$ that are also visible in $\mI_t$, according to the backward-to-forward flow difference $\Delta \mW_{t}^{f\leftarrow b}(\vx) = \mW_t^f + \mathcal{F}(\vx, \mW_t^f|\mW_t^b)$, similarly as the conditions~\cite{yin2018geonet,Meister:2018:UUL} that
\begin{multline}
\| \Delta\mW_t^{f\leftarrow b}(\vx) \|_1 < \\ \max\{\alpha, \beta (\| \mW_t^f(\vx) \|_1 + \| \mathcal{F}(\vx, \mW_t^f|\mW_t^b) \|_1)\}. \label{eq:occlusion_detection}
\end{multline}
Similarly, we also obtain the visibility mask $\mM_{t\leftarrow0}$ about pixels in $\mI_t$ that are also visible in $\mI_0$, based on the same condition for the forward-to-backward flow difference $\Delta \mW_t^{b\leftarrow f}$.
The hyper-parameters are set as $\alpha=1.0, \beta=0.1$ in our experiments.
Note that $\mM_{t\leftarrow0}$ corresponds to the target frame $\mI_t$ and $\mM_{0\leftarrow t}$ is with the reference frame $\mI_0$.

\vspace{+1mm}
\noindent\textbf{Cycle-consistent Flow Learning~}
The bi-directional flows are learned to enforce their internal cycle consistency, where the valid (\ie, in non-occluded regions) forward (or backward) flows pointing from $\mI_0$ (or $\mI_t$) to $\mI_t$ (or $\mI_0$) are mirrored by the backward (forward) flows from the warped locations in $\mI_t$ (or $\mI_0$) to the original locations in $\mI_0$ (or $\mI_t$).
We define the cycle consistency objective $L_\text{cc}$ by $\ell_1$ norm as
\begin{multline}
L_\text{cc} = \sum_{t\in\mathcal{T}}\sum_\mathbf{x} \mM_{0\leftarrow t}(\vx) \cdot \| \mathbf{W}_t^f(\mathbf{x}) + \mathcal{F}(\vx, \mW_t^f|\mW_t^b)  \|_1 \\ + \mM_{t\leftarrow 0}(\vx)\cdot \| \mathbf{W}_t^b(\mathbf{x}) + \mathcal{F}(\vx, \mW_t^b|\mW_t^f)  \|_1. \label{eq:cross_consistency}
\end{multline}
The learned flows should also be constrained by the bi-directional photometric consistency in the valid regions as
\begin{multline}
L_\text{bi-vc} = \sum_{t\in\mathcal{T}}\sum_\mathbf{x} \mM_{0\leftarrow t}(\vx) \cdot \| \mI_0(\vx) - \mathcal{F}(\vx, \mW_t^f|\mI_t) \|_1 \\ + \mM_{t\leftarrow 0}(\vx)\cdot \| \mI_t(\vx) - \mathcal{F}(\vx, \mW_t^b|\mI_0)  \|_1. \label{eq:bidirectional_photometric_consistency}
\end{multline}
The bi-directional flows in the occlusion regions are otherwise na\"ively guessed with a smoothness prior within their neighborhood, as we only apply the non-occluded flows to generate the warped frames, as Eq.~\eqref{eq:backward_warping}, while leaving the occluded regions undefined.
%
% These initial synthesized sequences are inputted to our occlusion-aware image synthesis module for the final video generation.

\begin{figure}[t]
\centering
\includegraphics[width=1\linewidth]{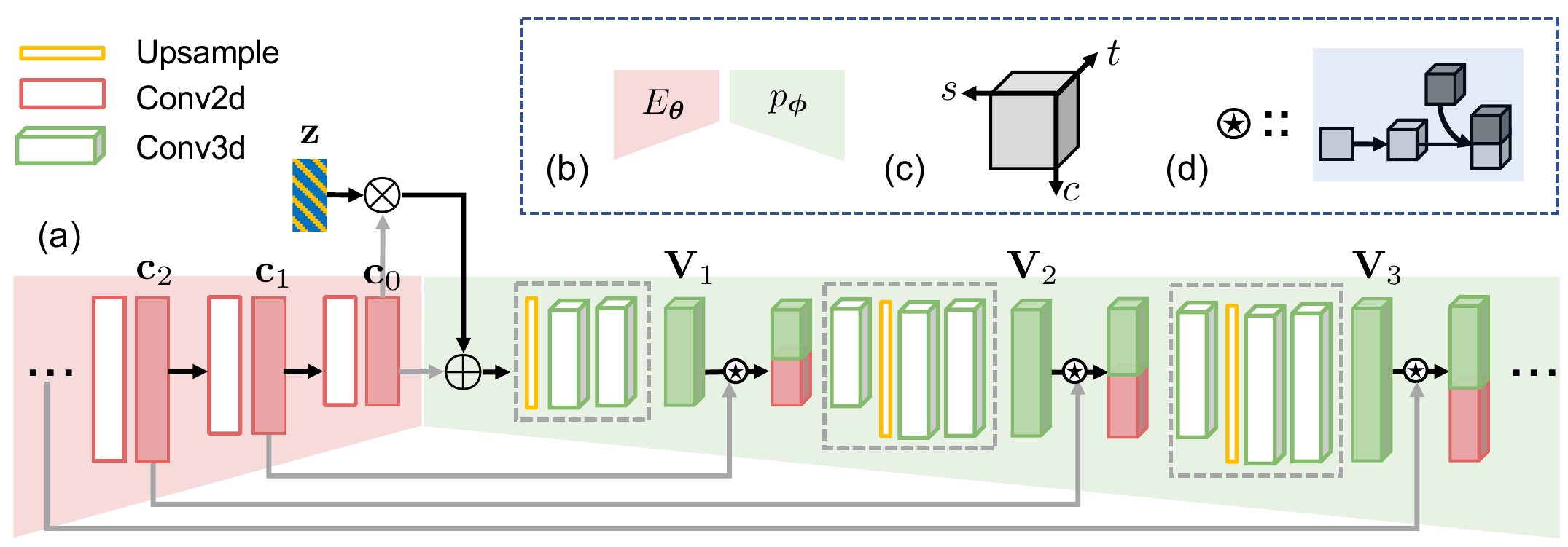}
\caption{(a) The compositional fusion for multi-level content-aware motion representation. (b) The pink color indicates the image encoder and the green color shows the flow generator. (c) The 3D feature volume, where $t, c$ and $s$ shows the time, channel and space dimension. (d) The 2D-to-3D fusion operation.}
\label{fig:content_aware_motion_distributions}
\end{figure}

\vspace{+1mm}
\noindent\textbf{Compositional Condition Fusion~}
In addition to cycle consistency, the proposed bi-directional flow generator $p_{\boldsymbol\phi}(\mathcal{W}_\mathcal{T}^f, \mathcal{W}_\mathcal{T}^b|\mathbf{z}, \mathbf{I}_0)$ are required to generate content-aware flows that are semantically corresponded to the content structures in $\mI_0$.
It is achieved by introducing a compositional condition fusion scheme into the main branch of bi-directional flow generator, which looks like the Hourglass structure~\cite{newell2016stacked} that gradually adapts the sampled motion features with multi-level content features $\{\vc_m\}_{m=1}^M$ extracted from the image encoder $E_{\boldsymbol\theta}(\mI_0)$, as illustrated in Fig.~\ref{fig:content_aware_motion_distributions}.

Our flow generator starts from fusing the sampled random variable $\vz$ with the top-level content feature vector $\vc_1$, by treating the content features as a depth-wise convolution kernel.
The fused motion features are upscaled by a deconvolution layer consisting of an upsampling operation and a 3D convolution layer.
%
 % the motion encoder $q_{\boldsymbol\psi}(\mathbf{z}|\mathbf{I}_0, \mathcal{I_T})$ and
%
3D convolution layers are employed throughout the main branch of $p_{\boldsymbol\phi}(\mathcal{W}_\mathcal{T}^f, \mathcal{W}_\mathcal{T}^b|\mathbf{z}, \mathbf{I}_0)$ to learn spatiotemporal features and offer more complex motion patterns in the generated flows.

The subsequent network repeats several stacked network blocks composed by a 2D-to-3D fusion block, an aforementioned deconvolution layer and an additional 3D convolution layer, before split into two branches of forward and backward flow subnets, as shown in Fig.~\ref{fig:content_aware_motion_distributions}(a).
The proposed 2D-to-3D fusion blocks fuses a content feature map $\vc_m$ and a corresponding 3D motion feature volume $\mV_m$, by at first concatenating $\vc_m$ along the channel axis of each time slice of $\mV_m$, and then being convoluted by one 3D convolution layer for a seamless feature fusion.
It suggests that the motion features in any time stamp should explicitly share the same content features with each other, similar as~\cite{tulyakov2017mocogan}.
%
% The concatenated feature volume is then convoluted by one 3D convolution layer to incorporate the content features seamlessly into the motion feature volume.

\begin{figure*}[t]
\centering
\includegraphics[width=0.9\linewidth]{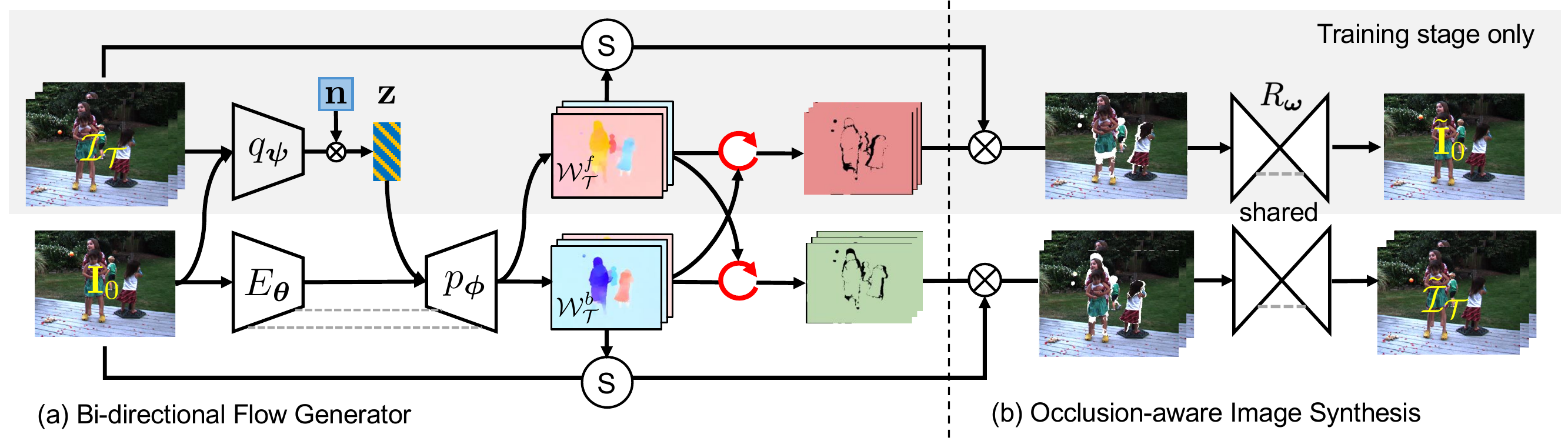}
\caption{The complete framework of our ImagineFlow model. The flow generator also includes a motion encoder $q_{\vpsi}$ to fulfill a CVAE~\cite{sohn2015learning} paradigm. The raw warped images together with their visibility masks are concatenated into the occlusion-aware image synthesis module. The area marked by a gray box indicates the modules used in training stage only. Best viewed on screen.}
\label{fig:framework}
\end{figure*}

\subsection{Occlusion-aware Image Synthesis}
\label{sub:occlusion_aware_image_synthesis}

Backward bilinear warping operation $\mathcal{F}(\vx, \mW_t^b|\mI_0)$ inherently suffers from warping artifacts, thus it could not produce visually plausible novel videos.
Li~\etal~\cite{li2018flow} applied an image refinement module to remove any warping artifacts.
We argue that explicit occlusion handling would be more effective in removing artifacts and inpainting contents in the occluded regions.

Unlike the frame interpolation studies~\cite{liu2017video,jiang2017super} that require at least two frames to infer occluded regions,
our method only needs one reference image $\mI_0$ to simultaneously infer bi-directional flows $\{ \mathcal{W}_\mathcal{T}^f, \mathcal{W}_\mathcal{T}^b \}$.
These flows infer the visibility masks $\{\mM_{0\leftarrow t}, \mM_{t\leftarrow0}\}_{t\in\mathcal{T}}$, according to Eq.~\eqref{eq:occlusion_detection}.
Consequently, we propose an occlusion-aware image synthesis module $R_{\vomega}$ that accepts the visibility mask $\mM_{t\leftarrow0}$ and the na\"ively warped frame $\mI_{t\leftarrow0}$.
It outputs a refined novel frame $\tilde{\mI}_{t\leftarrow0}$ with the suppression of warping artifacts.

The network for the image synthesis module is similar as those for the inpainting task~\cite{liu2018image}, which applies multi-level skip connections in an autoencoder.
The encoder borrows the same architecture of the VGG-19 up to the $\mathtt{ReLU4\_1}$ layer, while the decoder is symmetrical to the encoder with the nearest neighbor upsampling operations to replace the max pooling operations.
The skip connections link the $\mathtt{ReLU}k\_\mathtt{1}, k = \{1, 2, 3\}$ in the encoder to corresponding layers in the decoder.

In the training stage, the proposed synthesis module also refines the na\"ively warped frame $\mI_{0\leftarrow t}$ from the target frame $\mI_t$ to the reference frame $\mI_0$ with the help of the visibility mask $\mM_{0\leftarrow t}$.
We train this network using a perceptual loss~\cite{johnson2016perceptual} for bi-directional image synthesis:
\begin{multline}
L_\text{pp} = \sum_{t\in\mathcal{T}} \|\tilde{\mI}_{t\leftarrow0} - \mI_t \|_2^2 + \lambda\sum_{k=1}^5 \| \boldsymbol\Phi_k(\tilde{\mI}_{t\leftarrow0}) - \boldsymbol\Phi_k(\mI_t)\|_2^2 \\ 
+ \| \tilde{\mI}_{0\leftarrow t} - \mI_0 \|_2^2 + \lambda \sum_{k=1}^5 \| \boldsymbol\Phi_k(\tilde{\mI}_{0\leftarrow t}) - \boldsymbol\Phi_k (\mI_{0}) \|_2^2,
\end{multline}
where $\boldsymbol\Phi_k(\cdot)$ denote the features of a pretrained VGG-19 network at the layer $\mathtt{ReLU}k\_\mathtt{1}$, and $\lambda$ is to balance the terms.

The proposed occlusion-aware image synthesis module is able to fill unreliable regions with semantically meaningful contents, and hallucinate fine details to overcome the blurring artifacts caused by the bilinear warping operation.
This model is learned in an unsupervised fashion and can be incorporated with the aforementioned bi-directional flow generator for an end-to-end system for flow generation and image synthesis.

\subsection{Network Training and Implementation}
\label{sub:network_structure}

We apply the CVAE~\cite{sohn2015learning} to learn our conditioned generative model.
The complete system includes
(1) \emph{Motion Encoder} $q_{\boldsymbol\psi}(\mathbf{z}|\mathbf{I}_0, \mathcal{I_T})$ that produces a latent motion prior distribution encoding the underlying motions for the input sequences $\mathcal{I_T}$ \wrt the reference frame $\mathbf{I}_0$; 
(2) \emph{Image Encoder} $E_{\boldsymbol\theta}(\mathbf{I}_0)$ that extracts content features of $\mathbf{I}_0$ and it is included in the bi-directional flow generator;
(3) \emph{Bi-directional Flow Generator} $p_{\boldsymbol\phi}(\cW_\cT^f, \mathcal{W}_\mathcal{T}^b|\mathbf{z}, \mathbf{I}_0)$ that generates bi-directional flow fields from sampled motion variables $\mathbf{z}$ and correlated with the content features in $E_{\boldsymbol\theta}(\mathbf{I}_0)$;
(4) \emph{Occlusion-aware Image Synthesis} $R_{\vomega}(\cdot)$ that synthesizes the final sequences based on a combined warped frames and their visibility masks.

\vspace{0.1cm}
\noindent\textbf{Training Phase.}
In the training phase, the motion encoder encodes a stack of adjacent frames $\{\mathcal{I_T}, \mathbf{I}_0\}$ as a 3D volume and produces mean and variance vectors to model the posterior $q_{\boldsymbol\psi}(\mathbf{z}|\mathbf{I}_0, \mathcal{I_T})$.
The image encoder $E_{\boldsymbol\theta}(\mI_0)$ extracts multi-level content features $\{\mathbf{c}_m\}_{m=1}^M$.
The bi-directional flow generator $p_{\boldsymbol\phi}(\mathcal{W}_\mathcal{T}^f, \mathcal{W}_\mathcal{T}^b|\mathbf{z}, \mathbf{I}_0)$ uses the sampled motion variables $\mathbf{z}$ from $q_{\boldsymbol\psi}(\mathbf{z}|\mathbf{I}_0, \mathcal{I_T})$ as the input.
At the end of the flow generator, we produce the initial bi-directional warped frames $\{\mI_{t\leftarrow0}, \mI_{0\leftarrow t}\}_{t\in\mathcal{T}}$ and their visibility masks $\{\mM_{t\leftarrow0}, \mM_{0\leftarrow t}\}_{t\in\mathcal{T}}$.
They are then inputted into the proposed occlusion-aware image synthesis module $R_{\boldsymbol\omega}(\cdot)$ to obtain the final synthesized frames $\{\tilde{\mI}_{t\leftarrow0}, \tilde{\mI}_{0\leftarrow t}\}_{t\in\mathcal{T}}$.

\begin{figure*}[t]
\centering
\includegraphics[width=\linewidth]{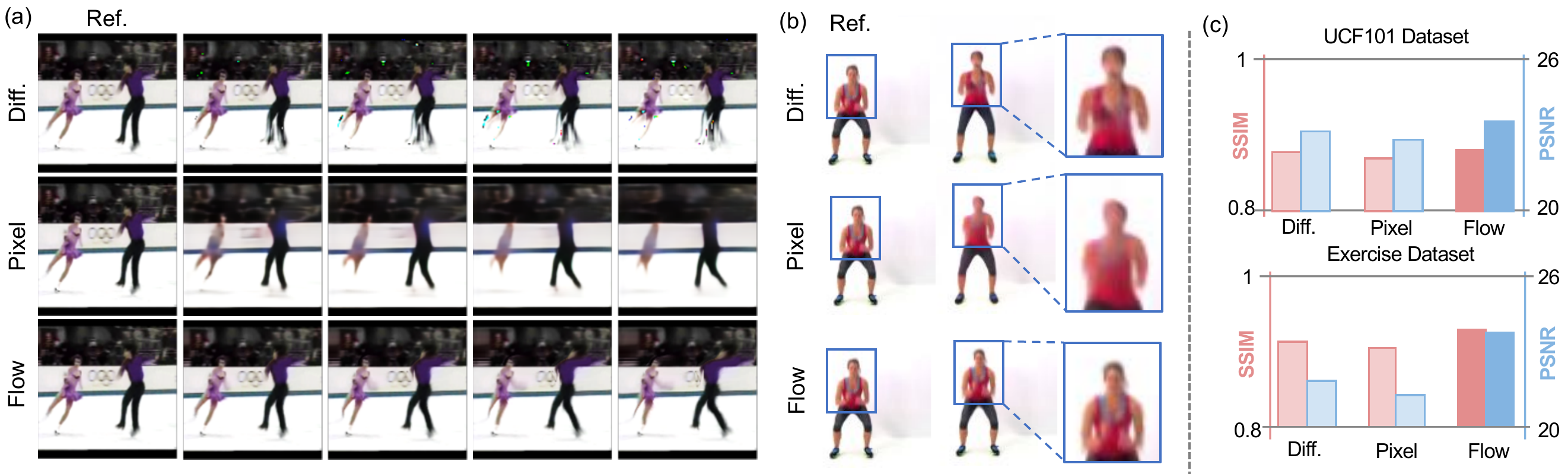}
\caption{Synthesizing novel frames by different motion representations. The leftmost image is the reference frame. (a) Sampled $4$-frame sequences in the UCF-101 dataset, and (b) sampled $1$-frame results in the Exercise dataset. (c) shows the PSNR@100 and SSIM@100 scores for both datasets.}
\label{fig:ablation_study_motion_representation}
\end{figure*}

The objective of our {ImagineFlow} model (Fig.~\ref{fig:framework}) extends the variational upper bound~\cite{kingma2013auto} of the CVAE model by adding the aforementioned losses
\begin{multline*}
L_{\boldsymbol\phi, \boldsymbol\psi, \boldsymbol\theta, \boldsymbol\omega}(\mathbf{I}_0, \mathcal{I_T}) = -D_\text{KL}[q_{\boldsymbol\psi}(\mathbf{z}|\mathbf{I}_0, \mathcal{I_T}) || \mathcal{N}(\mathbf{z}|\mathbf{0}, \mathbb{I})] + \\
\frac{1}{S}\sum_{s=1}^S \lambda_\text{bi-vc} L_\text{bi-vc} (\mathbf{z}^{(s)}) + \lambda_\text{cc} L_\text{cc} (\mathbf{z}^{(s)}) + \lambda_\text{pp} L_\text{pp}(\mathbf{z}^{(s)}),
\end{multline*}
where the bi-directional photometric consistency $L_\text{bi-vc}$, cycle flow consistency $L_\text{cc}$ and the perceptual loss $L_\text{pp}$ are monte-carlo integrated to serve as the negative log-likelihood for this generation model.
The KL-divergence aims at constraining the discrepancy between the posterior $q_{\boldsymbol\psi}(\mathbf{z}|\mathbf{I}_0, \mathcal{I_T})$ and the na\"ive motion prior. 
In addition to the above objective, we add a small amount of TV-$\ell_1$ norm to enhance the smoothness of the final images and flows.
$\lambda_\text{bi-vc} = 1.0, \lambda_\text{cc} = 0.05$ and $\lambda_\text{pp} = 1.0$.

\vspace{0.1cm}
\noindent\textbf{Test Phase.}
In the testing phase, we just require a plain motion prior $p(\mathbf{z}) = \mathcal{N}(\mathbf{z}|\mathbf{0}, \mathbb{I})$ to replace $q_{\boldsymbol\psi}(\mathbf{z}|\mathbf{I}_0, \mathcal{I_T})$ for the sampling of motion variable $\mathbf{z}$.
We just employ the backward flows $\mathcal{W}^b_\mathcal{T}$ and their visibility masks $\{ \mM_{t\leftarrow 0} \}_{t\in\mathcal{T}}$ for the final sequence generation.
The novel sequence $\{\tilde{\mI}_{t\leftarrow0}\}_{t\in\mathcal{T}}$ is generated by inputting $\{\mI_{t\leftarrow0}, \mM_{t\leftarrow0}\}_{t\in\mathcal{T}}$ into the occlusion-aware image synthesis module.

\section{Experiments}
\label{sec:evaluations_and_discussions}

\subsection{Settings}
\label{sub:system setup}

\noindent\textbf{Datasets.}
The proposed model is trained and evaluated on three popular video datasets: UCF-101 dataset~\cite{soomro2012ucf101}, Moving MNIST dataset~\cite{srivastava2015unsupervised} and Exercises dataset~\cite{xue2016visual}.
UCF-101 contains $13,320$ real video clips from $101$ action classes with substantial background movements.
Moving MNIST dataset is a synthetic dataset constructed by warping the digits in MINST dataset~\cite{lecun1998gradient} with affine transformations.
The Exercises dataset includes around $60$k pairs of frames from real workout videos with a static background.

\vspace{0.1cm}
\noindent\textbf{Implementation Details.}
\label{ssub:implementation_details}
Our system was implemented in PyTorch.
It is end-to-end trained by Adam optimizer, with a small learning rate of $0.001$, $\beta_1 = 0.9$ and $\beta_2 = 0.999$.
The batch size is $32$ and the train/test images are cropped and resized to $64\times64$ for the Moving MNIST dataset and $128\times128$ for the UCF-101 and Exercise Datasets.

We use two settings to show the superiority of the proposed method.
(1) For long-term video generation centered around a single frame, we set the number of rendering frames to $8$.
(2) To compare with prior arts on long-term and short-term future frame predictions, we modify our network by predicting the next $4$ frames and the next $1$ frame, accordingly.
Moreover, the $\mathtt{baseline}$ model copies the main structure but only preserves the backward flow branch and the highest fusion block, and removes the occlusion-aware image synthesis module.

\vspace{0.1cm}
\noindent\textbf{Evaluation Metrics.}
\label{ssub:evaluation_metrics}
We also quantitatively evaluate the models in addition to subjective tests.
We sample $100$ sequences for one reference frame and report the best PSNR and SSIM~\cite{wang2004image}, named as PSNR@100 and SSIM@100.
A good model should synthesize at least one sequence that is similar to the ground-truth.
We also apply the recent Frechet Inception Distance (FID) based on I3D model to evaluate the perceptual performance of the generated sequences.

\subsection{Ablation Study}
\label{sub:ablation_study}

The unique advantage of the proposed ImagineFlow model is its capability of learning robust flow distributions and synthesizing visually plausible videos.
It includes several pivotal components that contribute to this feature.

\begin{figure}[t]
\centering
\includegraphics[width=1.0\linewidth]{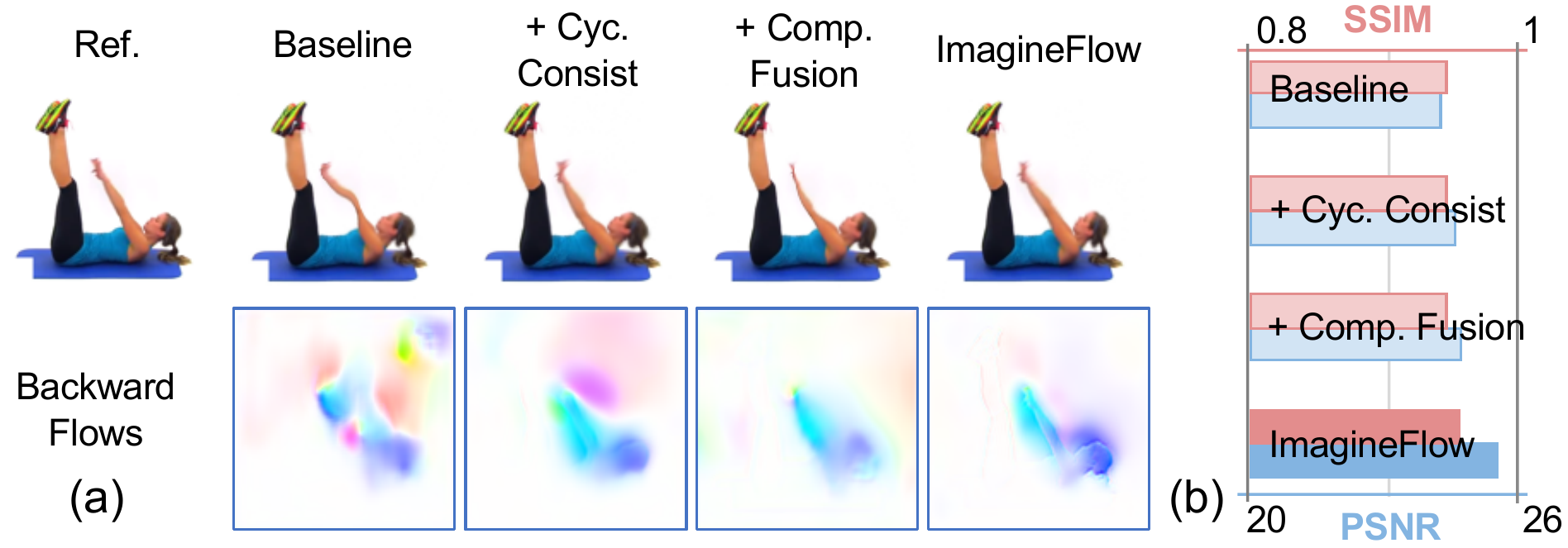}
\caption{Component comparison of the bi-directional flow generator. (a) are predictions of one reference image in the Exercise dataset. Each result is shown with its aligned backward flow field. (b) PSNR@100 and SSIM@100 scores. Best viewed on screen.}
\label{fig:ablation_study_component_wise_comparison}
\end{figure}

% \subsubsection{Motion Representation}
% \label{ssub:motion_representation}
\vspace{+1mm}
\noindent\textbf{Motion Representation.}
We show that flows are more reliable than difference images and pixel values in synthesizing novel frames.
To verify it, we add two more baselines by changing the output of our baseline network to RGB pixel values and difference images (named as $\mathtt{Flow}$, $\mathtt{Pixel}$ and $\mathtt{Diff}$, respectively).
As shown in Fig~\ref{fig:ablation_study_motion_representation}(a), in the exercise dataset, $\mathtt{Diff}$ blurs out the face details, but $\mathtt{Pixel}$ is often too blur to capture upper torsos and arms.
In contrast, $\mathtt{Flow}$ preserves the detailed contents and produces fewer visual artifacts around body parts.
In addition, in the UCF-101 dataset (Fig.~\ref{fig:ablation_study_motion_representation}(b)), $\mathtt{Pixel}$ fails to capture large motions and $\mathtt{Diff}$ produces severe aliasing around the legs of the athletes.
%
% \texttt{base-flow} generates faithful ice-dancing moves from the captured athletes. 
%
% It is difficult to describe motions as difference images or raw pixel values since it is usually harder for a network to learn unconstrained pixel changes or raw pixel values than well distributed flow fields.
%
Evaluations in Fig.~\ref{fig:ablation_study_motion_representation}(c) also quantitatively prove that the flow-based baseline outperforms the rest counterparts.
%
% Moreover, as indicated in the experimental comparison in Fig.~\ref{fig:comparisons}, our complete framework produces a much better photo-realistic rendering of novel sequences in terms of quantitative and qualitative evaluations, against the state-of-the-art flow-agnostic methods.

% \subsubsection{Component-wise Comparison}
% \label{ssub:component_wise_comparison}

\vspace{+1mm}
\noindent\textbf{Component-wise Comparison.}
Firstly, we validate the bi-directional flow generator.
$\mathtt{Cyc.~Consist}$ outputs bi-directional flows and applies cycle consistent flow learning, while $\mathtt{Comp.~Fusion}$ adds skip connections between the image encoder and flow decoder.
As shown in Fig.~\ref{fig:ablation_study_component_wise_comparison}(a), compared to the baseline model, $\mathtt{Cyc.~Consist}$ encourages more physically consistent flows around the upper body so that there are much fewer visual distortions in synthesizing arms.
But the generated flows are blur and cannot capture with the content well.
$\mathtt{Comp.~Fusion}$ applies multi-level content features to regularize the spatial structure of the predicted flows (\eg, small flows in the background and homogeneous flows aligned with the upper body).
But this structure does not generate physically reasonable flows, so the synthesized arms suffer from warping artifacts (\eg, the arms are much thinner).
Their combination (\ie, the flow generator in the ImagineFlow model) performs the best and the predicted flows are not only physically reliable but also consistent with the captured contents.
Apart from the qualitative results, either PSNR @100 or SSIM@100 reports similar performance gains in Fig.~\ref{fig:ablation_study_component_wise_comparison}(c), which suggest that the proposed components are complementary to each other.

\begin{figure}
\centering
\includegraphics[width=\linewidth]{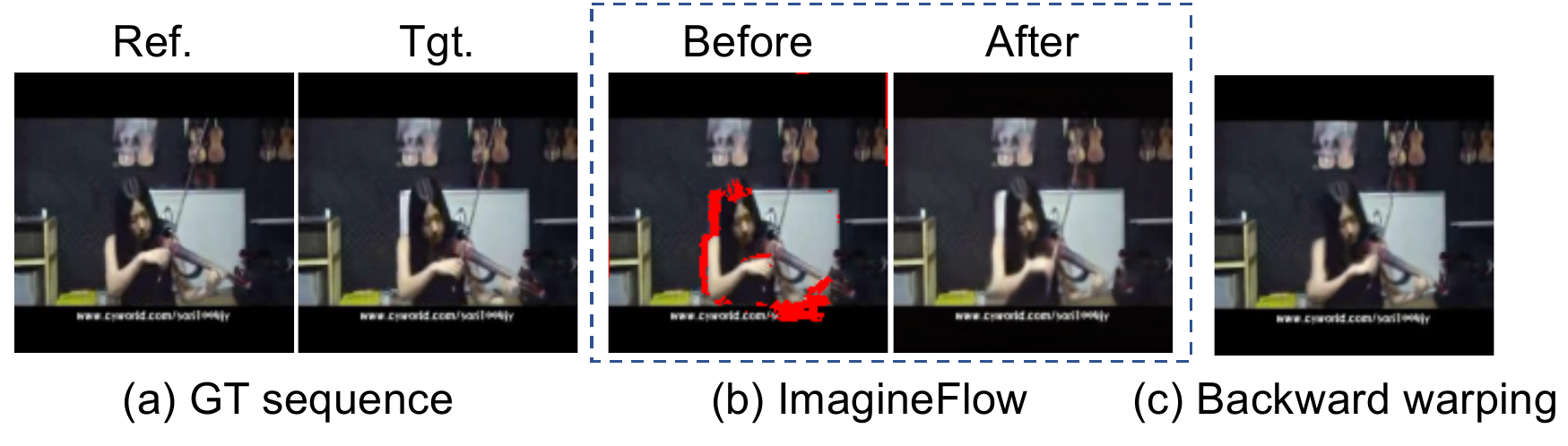}
\caption{Frame synthesis by the flow-based frame synthesis module. (a) Groundtruth reference and target frames. (b) \emph{Left:} Initial occlusion-aware warped target frame, where occlusions are marked in red color; \emph{Right}: After the occlusion-aware image synthesis module. (c) Na\"ive result by backward warping of the reference frame. Best viewed on screen.}
\label{fig:occlusion_handling}
\end{figure}

Our flow-based frame synthesis faithfully inpaints unreliable regions in the occlusion-aware warped frames.
For instance, the violinist in Fig.~\ref{fig:occlusion_handling}(a) is moving right, thus the occluded white board should be revealed in the target frame.
Our flow generator successfully discovers these unreliable regions (see Fig.~\ref{fig:occlusion_handling}(b)), and our frame synthesis module completes these regions and guesses the structure of the white board as what is desired.
As for comparison, we also show the backward warped target frame in Fig.~\ref{fig:occlusion_handling}(c).
Since it renders novel frames solely based on pixels in the reference image, the occluded background cannot be discovered, even though the underlying flows are estimated accurately.

\begin{figure}[t]
\centering
\includegraphics[width=\linewidth]{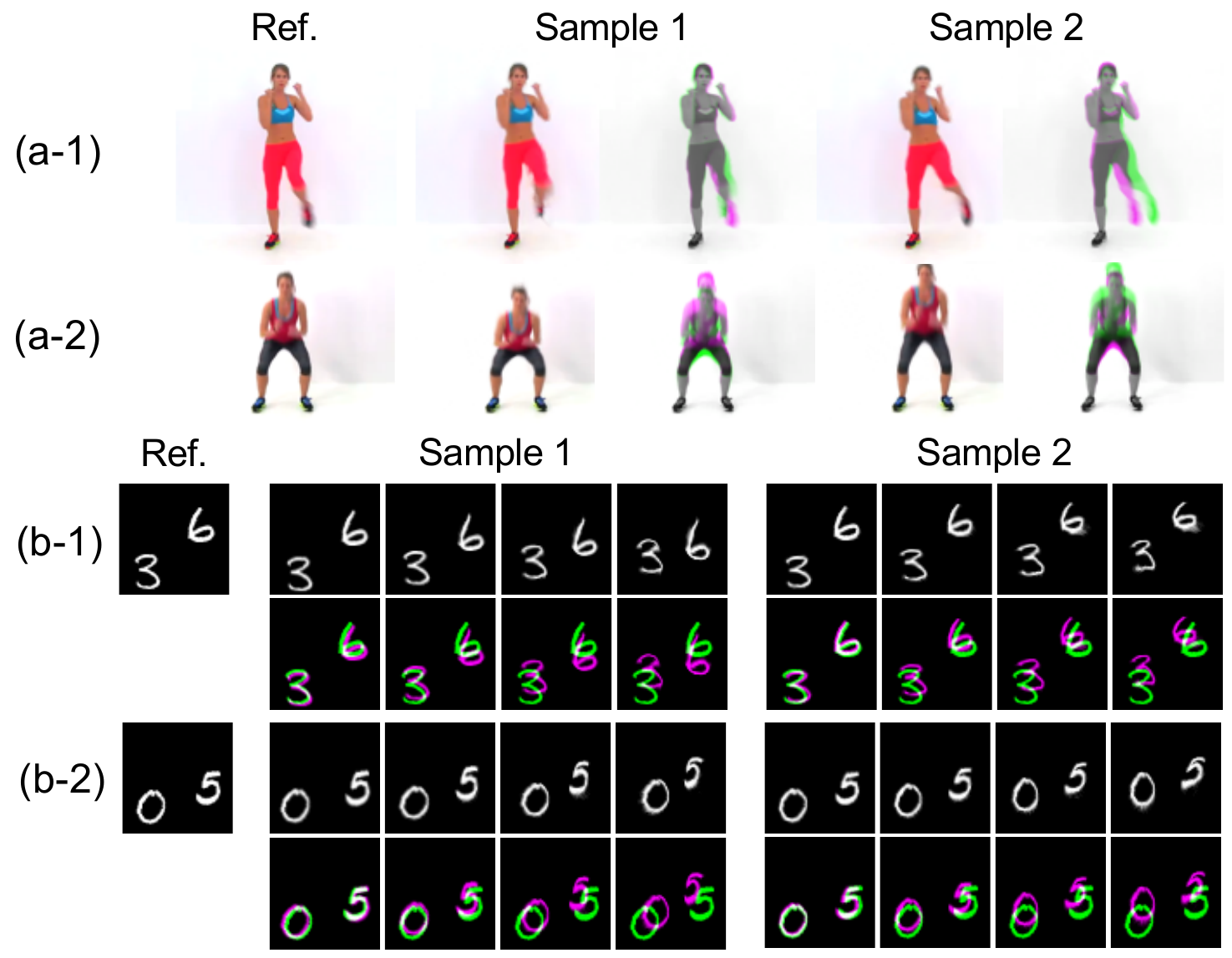}
\caption{Motion diversity. (a) Sampled next frames in the Exercise dataset. (b) Sampled $4$-frame sequences in the Moving MNIST dataset. Motions are illustrated as a RGB image encoding overlapped frames. Best viewed on screen.}
\label{fig:ablation_study_motion_diversity}
\end{figure}

\begin{figure*}[t]
\centering
\includegraphics[width=\linewidth]{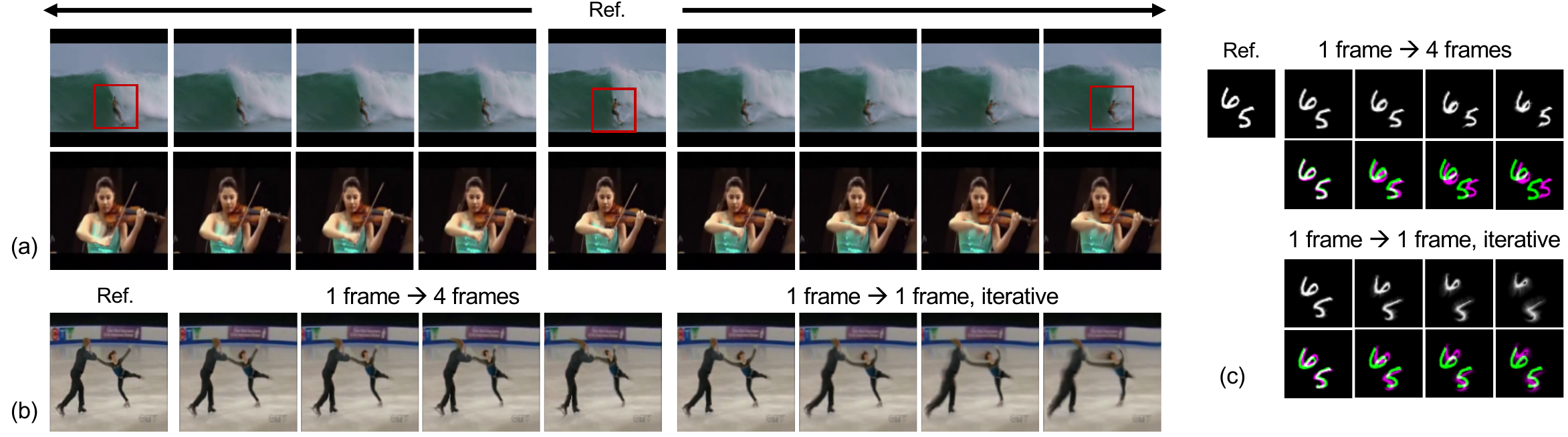}
\caption{(a) $9$-frame sequence generation on the UCF-101 dataset, where the center frame is the reference image. (b) and (c) compare the long-term ($1$ frame to $4$ frames prediction) and the short-term ($1$-frame to $1$-frame iterative prediction) ImagineFlow models in capturing complex motions and preserving structural coherence in the UCF-101 and Moving MNIST datasets. In (c) the color coded motions are illustrated for better visualization. Best viewed on screen.}
\label{fig:ablation_study_long_term_complex_motions}
\end{figure*}

% \begin{figure}[t]
% \centering
% \includegraphics[width=0.9\linewidth]{fig/fig_complex_motion_v2.pdf}
% \vspace{-3mm}
% \caption{The proposed long-term model predicts more complex motions than the iterative variant and preserves structural coherence of the contents. Best viewed on screen.}
% \label{fig:ablation_study_complex_motion_synthesis}
% \end{figure}

% \subsubsection{Motion Diversity}
% \label{ssub:motion_diversity}

\vspace{+1mm}
\noindent\textbf{Motion Diversity.}
The proposed method shows sufficient diversity in generating novel sequences.
In Fig.~\ref{fig:ablation_study_motion_diversity}, different samples for one reference frame in the Exercise and Moving MNIST datasets demonstrate diverse motion variations.
The motion is illustrated by creating a RGB image where the magenta channels are from the sampled frame and the green channel from the reference frame, as suggested in~\cite{xue2016visual}.
For examples, Fig.~\ref{fig:ablation_study_motion_diversity}(a-1) shows diversified motions around the legs and Fig.~\ref{fig:ablation_study_motion_diversity}(a-2) visualizes different squat actions.
The Moving MNIST dataset gives long-term motion patterns for two digits.
Sample 1 in Fig.~\ref{fig:ablation_study_motion_diversity}(b-1) shows contractive and clockwise motions, while sample 2 in Fig.~\ref{fig:ablation_study_motion_diversity}(b-2) depicts an ascending digit pair.

% \subsubsection{Motion Complexity}
% \label{ssub:capturing_complex_motions}

\vspace{+1mm}
\noindent\textbf{Motion Complexity.}
The proposed ImagineFlow model can capture complex and long-term motions.
In Fig.~\ref{fig:ablation_study_long_term_complex_motions}(a), we demonstrate sampled $9$-frame sequences given the $5^\text{th}$ frames as the reference.
The sequence ``surfing'' has varying tidal waves over time nearby the surfing board, and the athlete is blending over.
In the sequence ``violinist'', our ImagineFlow model covers occluded regions and preserves content structures with meaningful motions in playing violin.
The long-term ImagineFlow model samples more complex motion patterns than the short-term one, as shown in Fig.~\ref{fig:ablation_study_long_term_complex_motions}(b) and (c).
The example in Fig.~\ref{fig:ablation_study_long_term_complex_motions}(b) shows that the long-term prediction is capable of guessing the rotation of the female dancer while the iterative short-term prediction will gradually distort and blur the contents.
The example in Fig.~\ref{fig:ablation_study_long_term_complex_motions}(b) also finds that the iterative variant fails to preserve the spatial structure of the digits.

\begin{figure}[t]
\centering
\includegraphics[width=\linewidth]{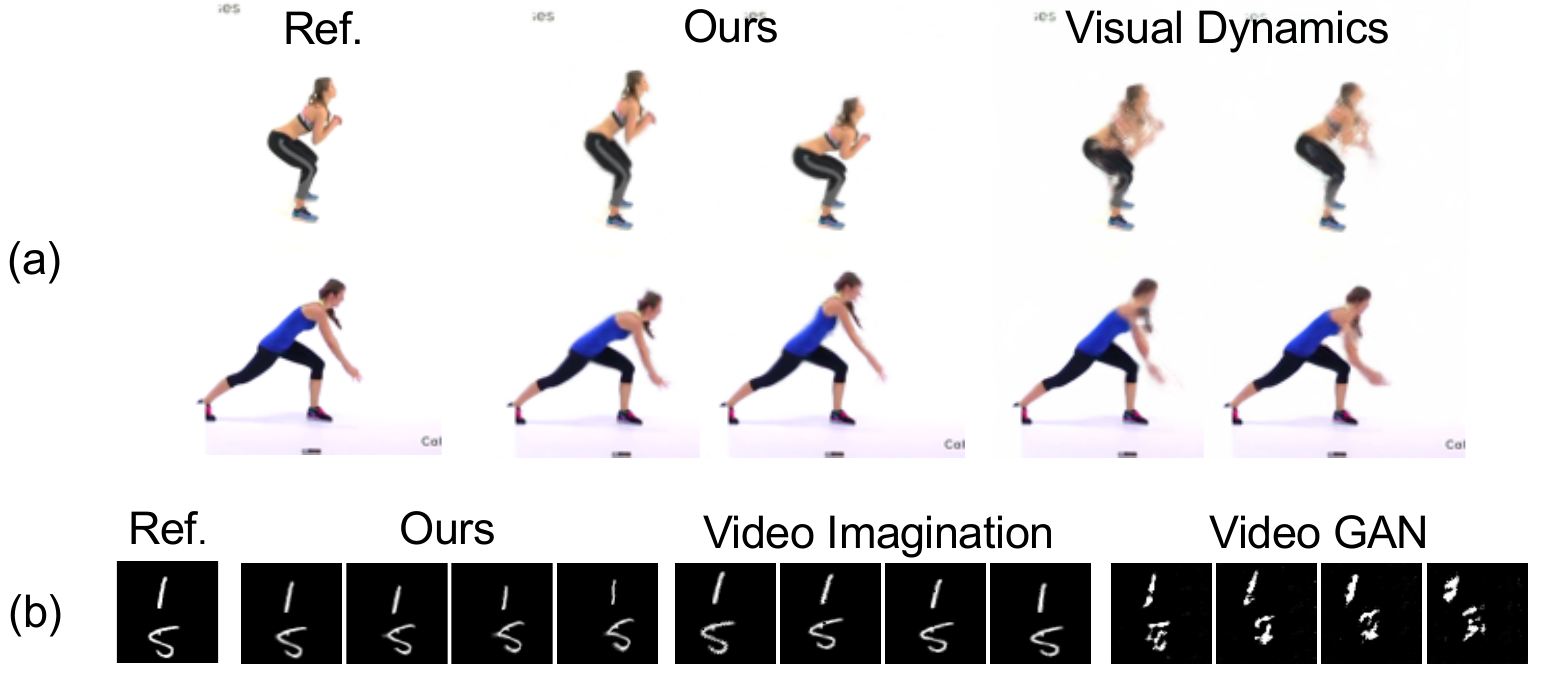}
\caption{Visual comparison with the state-of-the-art methods on the (a) Exercises dataset, (b) Moving MNIST dataset. }
\label{fig:comparisons}
\end{figure}

\subsection{Experimental Comparisons}
\label{sub:experimental_comparisons}

Our ImagineFlow model is compared with various single image based video synthesis methods such as Visual Dynamics~\cite{xue2016visual}, Video GAN~\cite{vondrick2016generating}, Video Imagination~\cite{chen2017video}, MoCoGAN~\cite{tulyakov2017mocogan} and Li~\etal~\cite{li2018flow}, in which Visual Dynamics is designed for the next frame prediction but Video GAN, Video Imagination, MoCoGAN and Li~\etal can be applied for longer video generation.

\emph{Exercise Dataset:}
As shown in Fig.~\ref{fig:comparisons}(a), the proposed model and Visual Dynamics are compared by sampling two future frames given a same reference frame.
Our results do not only present distinct motions but also preserve structural coherence without blurring and distortions.
However, Visual Dynamics, which depends on difference images as the motion representation, produced blurry future frames with mixed arm patterns (shown in the first row) or distorted hand structures (as in the second row).

\emph{Moving MNIST Dataset:}
Similar as our method, Video Imagination is able to render reliable predictions on the Moving MNIST dataset, as shown in Fig.~\ref{fig:comparisons}(b). 
It is because these digits are generated by affine transformations and thus their motions fit the mechanism of Video Imagination.
However, Video GAN produces large motions but the shapes of the digits are severely distorted.

\emph{UCF-101 Dataset:}
Fig.~\ref{fig:comparisons}(c) shows the visual comparison with Video Imagination, Video GAN and Li~\etal on the \emph{ice dancing} sequence about synthesizing $4$-frame sequences.
These reference GAN based methods, such as Video GAN and MoCoGAN usually fail when the videos contain over-complex spatio-temporal patterns (\ie, \emph{ice dancing}) that either the motion cannot be well captured or the novel contents will be distorted tremendously. 
On the other hand, Video Imagination produces noisy future contents with aliasing artifacts from multiple warped contents from the reference frame.
Li~\etal~\cite{li2018flow} also employ the flows as the medium for motion representation, but the predicted motions are too subtle and the rendered novel frames are distorted out of the space of real images.
Our prediction is able to render photorealistic ``ice dancing'' moves of the athletes without structural distortions.

\begin{figure}[t]
\centering
\includegraphics[width=\linewidth]{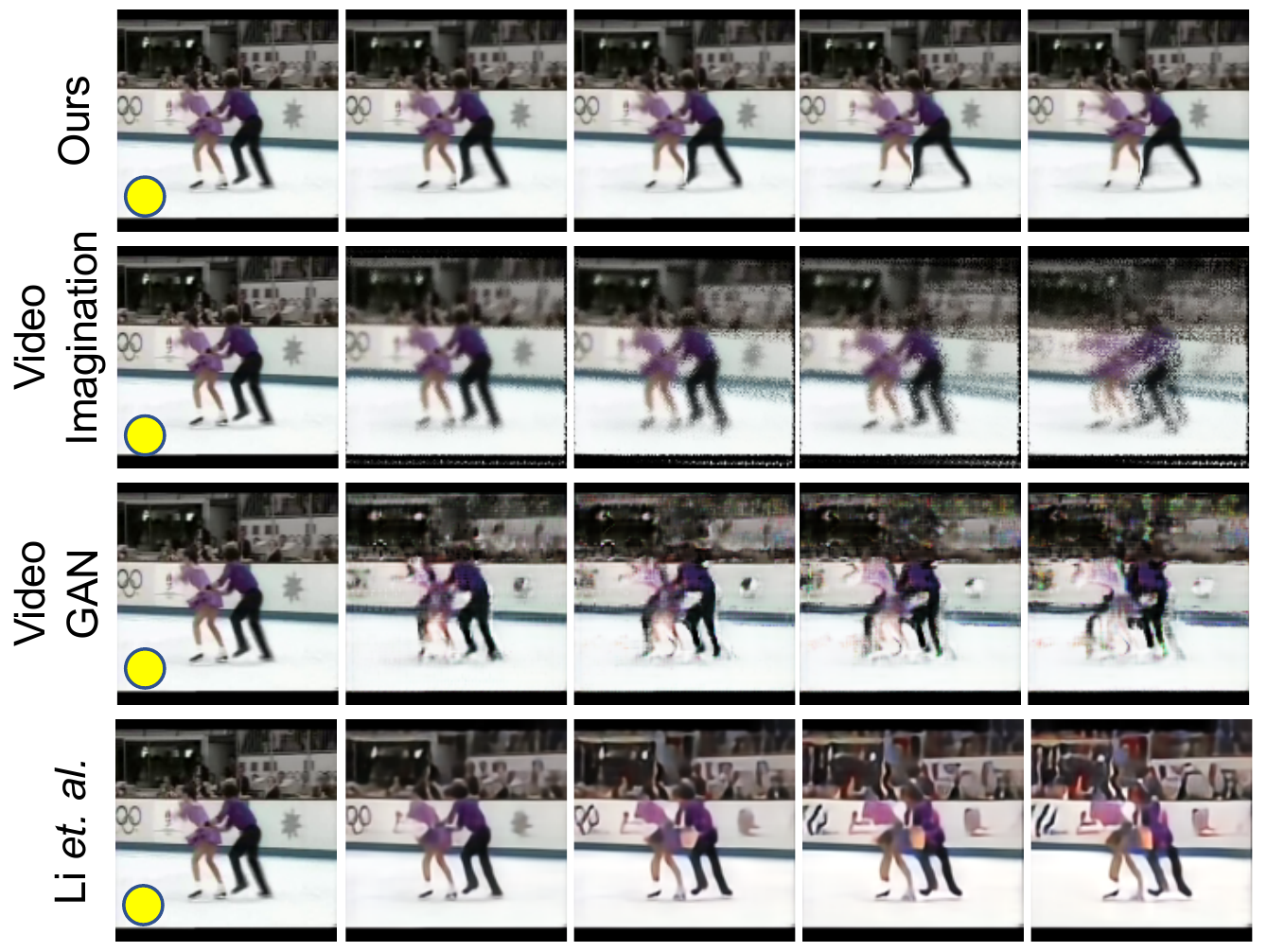}
\caption{Visual comparison with the state-of-the-art methods on the UCF-101 dataset. The start frame is marked by yellow cycle.}
\vspace{-3mm}
\label{fig:comparisons}
\end{figure}

\begin{table}
\small
\centering
\begin{tabular}{c|c|c|c}
\hline
\hline
& MoCoGAN~\cite{tulyakov2017mocogan} & Li~\etal~\cite{li2018flow} & ImagineFlow \\
\hline
FID & 7.23 & 2.53 & 1.82 \\
\hline
\hline
\end{tabular}
\vspace{+1mm}
\caption{FID score comparison on \emph{ice dancing} sequence.}
\end{table}

For all three datasets, the proposed ImagineFlow model finds reliable motion patterns, and remarkably preserve the structure coherence in the rendered novel frames.
We show the FID score \vs MoCoGAN~\cite{tulyakov2017mocogan} and Li~\etal~\cite{li2018flow}, our method receives the best quantitative performance.
Moreover, although no adversarial training is introduced in our system, our model is still able to generate photo-realistic results with fine details.

\section{Conclusion}
\label{sec:conclusion}

In this paper, we propose a novel probabilistic framework that samples video sequences from a single image.
Our model improves the CVAE framework with a bi-directional flow generator and a compositional fusion structure, thus is able to learn content-aware and structural coherent flow distributions.
The involved flow-based frame synthesis module renders high-quality novel sequences and solves the rendering artifacts inherently in the warping-based operation.
We have shown that the proposed model performs well on both on synthetic and real-world videos.

\begin{table*}[t]
\centering
\begin{tabular}{M{2cm}|M{2cm}|M{1.8cm}|M{1.8cm}|M{1.5cm}|M{1.5cm}|M{1.5cm}}
\toprule
name & type & num filters & filter size & stride & padding & activation\\
\midrule
\texttt{conv1} & \texttt{conv3d} & $64$ & $3\times3\times3$ & $1\times1\times1$ & $1\times1\times1$ & \texttt{lrelu}\\
\texttt{pool1} & \texttt{maxpool3d} & - & - & $1\times2\times2$ & - & - \\
\texttt{conv2} & \texttt{conv3d} & $64$ & $3\times3\times3$ & $1\times1\times1$ & $1\times1\times1$ & \texttt{lrelu}\\
\texttt{pool2} & \texttt{maxpool3d} & - & - & $1\times2\times2$ & - & - \\
\texttt{conv3} & \texttt{conv3d} & $128$ & $3\times3\times3$ & $1\times1\times1$ & $1\times1\times1$ & \texttt{lrelu}\\
\texttt{pool3} & \texttt{maxpool3d} & - & - & $1\times2\times2$ & - & - \\
\texttt{conv4} & \texttt{conv3d} & $256$ & $3\times3\times3$ & $1\times1\times1$ & $1\times1\times1$ & \texttt{lrelu}\\
\texttt{pool4} & \texttt{maxpool3d} & - & - & $2\times2\times2$ & - & - \\
\texttt{conv5} & \texttt{conv3d} & $512$ & $3\times3\times3$ & $1\times1\times1$ & $1\times1\times1$ & \texttt{lrelu}\\
\texttt{pool5} & \texttt{maxpool3d} & - & - & $2\times2\times2$ & - & - \\
\midrule
$\boldsymbol\mu$ & \texttt{conv3d} & $1024$ & $2\times4\times4$ & $1\times1\times1$ & $0\times0\times0$ & \texttt{linear}\\
$\log \boldsymbol\sigma$ & \texttt{fc} & $1024$ & $2\times4\times4$ & $1\times1\times1$ & $0\times0\times0$ & \texttt{linear} \\
\bottomrule
\end{tabular}
\vspace{+1mm}
\caption{Network specification of the motion encoder.}
\label{tab:motion_encoder}
\end{table*}

\begin{table*}[t]
\centering
\begin{tabular}{M{1.5cm}|M{1.5cm}|M{1.8cm}|M{1.8cm}|M{1.5cm}|M{1.5cm}|M{1.5cm}}
\toprule
name & type & num filters & filter size & stride & padding & activation\\
\midrule
$\mathbf{c}_5$ & \texttt{conv2d} & $64$ & $4\times4$ & $2\times2$ & $1\times1$ & \texttt{lrelu}\\
$\mathbf{c}_4$ & \texttt{conv2d} & $64$ & $4\times4$ & $2\times2$ & $1\times1$ & \texttt{lrelu}\\
$\mathbf{c}_3$ & \texttt{conv2d} & $128$ & $4\times4$ & $2\times2$ & $1\times1$ & \texttt{lrelu}\\
$\mathbf{c}_2$ & \texttt{conv2d} & $256$ & $4\times4$ & $2\times2$ & $1\times1$ & \texttt{lrelu}\\
$\mathbf{c}_1$ & \texttt{conv2d} & $512$ & $4\times4$ & $2\times2$ & $1\times1$ & \texttt{lrelu}\\
$\mathbf{c}_0$ & \texttt{conv2d} & $1024$ & $4\times4$ & $1\times1$ & $0\times0$ & \texttt{lrelu}\\
\bottomrule
\end{tabular}
\vspace{+1mm}
\caption{Network specification of the image encoder.}
\label{tab:image_encoder}
\end{table*}

\section*{Appendix: Detailed Network Architecture}
\label{sec:detailed_network_architecture}

The complete network consists of (a) a motion encoder $q_{\boldsymbol\psi}(\vz|\mI_0, \mathcal{I}_\mathcal{T})$, (b) a bi-directional flow generator $p_{\boldsymbol\phi}(\mathcal{W}_\mathcal{T}^f, \mathcal{W})\mathcal{T}^b|\vz, \mI_0)$, (c) an image encoder $E_{\boldsymbol\theta}(\mI_0)$ and (d) an occlusion-aware synthesis module $R_{\vomega}(\cdot)$.
In the following subsections, we will specify the network architectures for each component.

Note that batch normalization is applied in the entire network. $\mathtt{Leaky~ReLU}$ with $\lambda=0.2$ is the activation function used in each convolution layer.
But $\mathtt{linear}$ activation function is applied in the output layers.
Input images are normalized in the range $[0, 1]$.
We only illustrate the network for the input size of $128\times128$ and sequence length of $8$.

\vspace{+2mm}
\noindent\textbf{Motion Encoder.}
The motion encoder uses an image volume of size $N\times9\times128\times128\times3$ as its input.
For a sample in one batch, it stacks an image sequence $\mathcal{I}_\mathcal{T}$ and its reference frame $\mI_0$ along the time dimension.
$N$ is the batch size and $|\mathcal{I_T}| = 8$.
The output has two branches, one indicates the means $\boldsymbol\mu$ and the other is $\log \boldsymbol\sigma$, as the parameters for the Gaussian posterior $q_{\boldsymbol\psi}(\vz|\mI_0, \mathcal{I}_\mathcal{T})$.
This network is shown in Tab.~\ref{tab:motion_encoder}.

\vspace{+2mm}
\noindent\textbf{Image Encoder.}
The image encoder takes the reference image $\mI_0$ as the input, and outputs intermediate content features $\{ \vc_m \}_{m=0}^5$ through six consecutive 2D convolution layers.
The highest level of the content features is $\vc_1$, which is a $N\times1\times1\times1\times1024$ tensor.
The network specification is shown in Tab.~\ref{tab:image_encoder}.

\vspace{+2mm}
\noindent\textbf{Bi-directional Flow Generator.}
The bi-directional flow generator starts from sampling the motion variables $\vz$ either by the reparameterization trick as $\vz = \boldsymbol\mu + \boldsymbol\sigma \circ \boldsymbol\epsilon$, where $\boldsymbol\epsilon \sim \mathcal{N}(\mathbf{0}, \mathbb{I})$, or the standard Gaussian distribution $\vz \sim \mathcal{N}(\mathbf{0}, \mathbb{I})$.

The sampled motion tensor of size $N\times1\times1\times1\times1024$ is then inputted into the flow generator.
The first fusion is similar as the depthwise convolution with the convolution kernel as the content feature $\vc_1$.
The subsequent fusions are operated as: at first concatenation of the motion features and the content features along the time dimension, and then a 3D convolution layer to fuse across all dimensions.
We use the nearest neighboring upsampling with a 3D convolution layer to deconvolute the preceding motion features.
The network outputs are flow volumes of size $N\times8\times128\times128\times4$, \ie, the number of bi-directional flows is $8$. The forward and backward flows are split along the last dimension.

The network architecture of the flow generator is shown in Tab.~\ref{tab:flow_generator}, its main branch actually mirrored the structure of the motion encoder.

\begin{table*}[t]
\centering
\begin{tabular}{M{1.5cm}|M{1.5cm}|M{1.8cm}|M{1.8cm}|M{1.5cm}|M{1.5cm}|M{1.5cm}}
\toprule
name & type & num filters & filter size & stride & padding & activation\\
\midrule
$\mV_0$ & \multicolumn{6}{c}{Sampling of $\{\boldsymbol\mu, \boldsymbol\sigma\}$ or na\"ive Gaussian prior} \\
\hline
\texttt{fusion0} & \multicolumn{6}{c}{Depthwise convolution of $\mV_0$ by the kernel from $\vc_1$} \\
\hline
$\mV_1$ & \texttt{deconv3d} & 512 & $2\times4\times4$ & $1\times1\times1$ & $0\times0\times0$ & \texttt{lrelu} \\
\hline
\multirow{2}{*}{\texttt{fusion1}} & \multicolumn{6}{c}{concatenate $\mathbf{V}_1$ with $\mathbf{c}_1$ along the channel for each time slice} \\
\cline{2-7}
& \texttt{conv3d} & $512$ & $3\times3\times3$ & $1\times1\times1$ & $1\times1\times1$ & \texttt{lrelu} \\
\hline
\multirow{2}{*}{\texttt{upconv1}} & \texttt{upsample} & - & - & $2\times2\times2$ & - & - \\
\cline{2-7}
& \texttt{conv3d} & $256$ & $3\times3\times3$ & $1\times1\times1$ & $1\times1\times1$ & \texttt{lrelu} \\
\hline
$\mathbf{V}_2$ & \texttt{conv3d} & $256$ & $3\times3\times3$ & $1\times1\times1$ & $1\times1\times1$ & \texttt{lrelu} \\
\hline
\multirow{2}{*}{\texttt{fusion2}} & \multicolumn{6}{c}{concatenate $\mathbf{V}_2$ with $\mathbf{c}_2$ along the channel for each time slice} \\
\cline{2-7}
& \texttt{conv2d} & $256$ & $3\times3\times3$ & $1\times1\times1$ & $1\times1\times1$ & \texttt{lrelu} \\
\hline
\multirow{2}{*}{\texttt{upconv2}} & \texttt{upsample} & -- & -- & $2\times2\times2$ & -- & -- \\
\cline{2-7}
 & \texttt{conv3d} & $128$ & $3\times3\times3$ & $1\times1\times1$ & $1\times1\times1$ & \texttt{lrelu}\\
\hline
 $\mathbf{V}_3$ & \texttt{conv3d} & $128$ & $3\times3\times3$ & $1\times1\times1$ & $1\times1\times1$ & \texttt{lrelu} \\
\hline
\multirow{2}{*}{\texttt{fusion3}} & \multicolumn{6}{c}{concatenate $\mathbf{V}_3$ with $\mathbf{c}_3$ along the channel for each time slice} \\
\cline{2-7}
& \texttt{conv3d} & $128$ & $3\times3\times3$ & $1\times1\times1$ & $1\times1\times1$ & \texttt{lrelu} \\
\hline
\multirow{2}{*}{\texttt{upconv3}} & \texttt{upsample} & -- & -- & $1\times2\times2$ & -- & -- \\
\cline{2-7}
 & \texttt{conv3d} & $64$ & $3\times3\times3$ & $1\times1\times1$ & $1\times1\times1$ & \texttt{lrelu}\\
\hline
$\mathbf{V}_4$ & \texttt{conv3d} & $64$ & $3\times3\times3$ & $1\times1\times1$ & $1\times1\times1$ & \texttt{lrelu} \\
\hline
\multirow{2}{*}{\texttt{fusion4}} & \multicolumn{6}{c}{concatenate $\mathbf{V}_4$ with $\mathbf{c}_4$ along the channel for each time slice} \\
\cline{2-7}
& \texttt{conv3d} & $64$ & $3\times3\times3$ & $1\times1\times1$ & $1\times1\times1$ & \texttt{lrelu} \\
\hline
\multirow{2}{*}{\texttt{upconv4}} & \texttt{upsample} & -- & -- & $1\times2\times2$ & -- & -- \\
\cline{2-7}
 & \texttt{conv3d} & $64$ & $3\times3\times3$ & $1\times1\times1$ & $1\times1\times1$ & \texttt{lrelu}\\
\hline
$\mathbf{V}_5$ & \texttt{conv3d} & $64$ & $3\times3\times3$ & $1\times1\times1$ & $1\times1\times1$ & \texttt{lrelu} \\
\hline
\multirow{2}{*}{\texttt{fusion5}} & \multicolumn{6}{c}{concatenate $\mathbf{V}_5$ with $\mathbf{c}_5$ along the channel for each time slice} \\
\cline{2-7}
& \texttt{conv3d} & $64$ & $3\times3\times3$ & $1\times1\times1$ & $1\times1\times1$ & \texttt{lrelu} \\
\hline
\multirow{2}{*}{\texttt{upconv5}} & \texttt{upsample} & -- & -- & $1\times2\times2$ & -- & -- \\
\cline{2-7}
 & \texttt{conv3d} & $64$ & $3\times3\times3$ & $1\times1\times1$ & $1\times1\times1$ & \texttt{lrelu}\\
\hline
\texttt{output} & \texttt{conv3d} & $4$ & $3\times3\times3$ & $1\times1\times1$ & $1\times1\times1$ & \texttt{linear} \\
\bottomrule
\vspace{+1mm}
\end{tabular}
\caption{Network specification of the bi-directional flow generator.}
\label{tab:flow_generator}
\end{table*}

\vspace{+2mm}
\noindent\textbf{Occlusion-aware Image Synthesis.}
The structure of the occlusion-aware frame synthesis is briefly depicted in the main article. It uses the network layers up to $\mathtt{conv4\_1}$ of VGG-19 as the encoder. Its decoder mirrors the encoder with nearest neighbor upsampling to replace the max pooling operation. The skip connections link encoding layers $\mathtt{conv}k\_\mathtt{1}, k = 1, 2, 3$ to their corresponding decoding layers.
The input of this network is a stacked tensor about the warped frame and its visibility map, along the channel dimension.

{\small
\bibliographystyle{ieee}
\bibliography{egbib}
}

\end{document}